\documentclass[pdflatex,sn-mathphys-num]{sn-jnl}% Math and Physical Sciences Numbered Reference Style
\usepackage{enumerate}
\usepackage{longtable}
\usepackage{graphicx}%
\usepackage{multirow}%
\usepackage{amsmath,amssymb,amsfonts}%
\usepackage{amsthm}%
\usepackage{mathrsfs}%
\usepackage[title]{appendix}%
\usepackage{xcolor}%
\usepackage{textcomp}%
\usepackage{manyfoot}%
\usepackage{booktabs}%
\usepackage{algorithm}%
\usepackage{algorithmicx}%
\usepackage{algpseudocode}%
\usepackage{listings}%
\usepackage{pdfpages}
\usepackage{subfigure}
\usepackage{tabularx}
\usepackage{color}
\usepackage{multirow}
\usepackage{booktabs}
\usepackage{graphicx}
\usepackage{amsmath}
\usepackage{color}
\usepackage[numbers,sort&compress]{natbib}
\usepackage[numbers]{natbib}

\theoremstyle{thmstyleone}%
\theoremstyle{thmstyletwo}%

\theoremstyle{thmstylethree}%

\begin{document}

\title[Article Title]{A Cascaded Information Interaction Network for Precise Image Segmentation}


\author[1]{\fnm{Hewen} \sur{Xiao}}\email{hoan.xiao@gmail.com}

\author*[2]{\fnm{Jie} \sur{Mei}}\email{jmei@hit.edu.cn}

\author[2]{\fnm{Guangfu} \sur{Ma}}\email{magf@hit.edu.cn}

\author[1]{\fnm{Weiren} \sur{Wu}}\email{wwrhitsz@163.com}
\affil[1]{\orgdiv{The Institute of Space Science and
Applied Technology}, \orgname{Harbin Institute of  Technology, Shenzhen}, \orgaddress{\postcode{518055}, \state{Guangdong}, \country{China}}}

\affil[2]{\orgdiv{School of Mechanical Engineering and Automation}, \orgname{Harbin Institute of  Technology, Shenzhen}, \orgaddress{\postcode{518055}, \state{Guangdong}, \country{China}}}


\abstract{
Visual perception plays a pivotal role in enabling autonomous behavior, offering a cost-effective and efficient alternative to complex multi-sensor systems. However, robust segmentation remains a challenge in complex scenarios. To address this, this paper proposes a cascaded convolutional neural network integrated with a novel Global Information Guidance Module. This module is designed to effectively fuse low-level texture details with high-level semantic features across multiple layers, thereby overcoming the inherent limitations of single-scale feature extraction. This architectural innovation significantly enhances segmentation accuracy, particularly in visually cluttered or blurred environments where traditional methods often fail. Experimental evaluations on benchmark image segmentation datasets demonstrate that the proposed framework achieves superior precision, outperforming existing state-of-the-art methods. The results highlight the effectiveness of the approach and its promising potential for deployment in practical robotic applications.
}




\maketitle

\section{Introduction}\label{sec1}

Among computer vision techniques, image segmentation plays a crucial role as an indispensable auxiliary technology. Its primary function is to decompose visual scenes into meaningful regions, thereby facilitating downstream tasks such as detection, tracking, and recognition. Improving segmentation quality directly enhances the robustness and accuracy of robotic perception systems, with extensive applications in obstacle avoidance~\cite{zhang2025safety}, navigation~\cite{liu2025autonomous}, and target tracking~\cite{roberts2012saliency}.

Traditional image segmentation methods rely heavily on hand-crafted features or intrinsic priors \cite{zhang2019synthesizing}, which often limit their adaptability in complex or cluttered scenes. Recent advances in deep learning, particularly Convolutional Neural Networks (CNNs), have significantly boosted segmentation performance by learning multi-level features from data~\cite{liu2019simple,GateNet2020,wang2019salient}. However, many CNN-based models~\cite{wang2018detect,wang2019iterative,wu2019cascaded} still struggle to balance fine-grained detail preservation and global contextual understanding due to the limitations of information interactions between multi-level features. Addressing this issue requires more effective multi-level and multi-scale representation mechanisms to enhance both spatial resolution and semantic abstraction.

To address challenges in visual perception in image segmentation, this paper introduces a cascaded neural network equipped with a global information guidance module, which effectively integrates low-level texture details and high-level semantic features across layers, overcoming the limitations of single-scale feature extraction. This design enhances segmentation accuracy, particularly in visually cluttered or blurred environments. We conducted extensive evaluations on standard image segmentation datasets to validate our approach. The results demonstrate that our method outperforms existing approaches in segmentation accuracy, highlighting its potential for real-time robotic applications in complex environments.

\section{Related Work}
Image segmentation aims to find regions of greatest interest to people in images. Traditional image segmentation approaches usually predict the saliency scores by utilizing hand-crafted cues or intrinsic priors~\cite{lee2016deep,pixel2021xu}. However, they are limited due to their low efficiency and bad generalization ability. With the rise of deep learning, recent methods mostly leverage convolutional neural networks (CNN) to make a pixel-to-pixel prediction.

Compared with traditional ones, CNN-based methods have shown superior performance on popular image segmentation benchmarks. Among them, early work \cite{wang2018detect,wang2019iterative,wu2019cascaded} mostly adopted an iterative or stage-wise manner to refine the predictions step by step. Some later methods \cite{liu2019simple,GateNet2020,chang2024deep} focus on designing new multi-scale feature-extracting modules and strategies based on the U-shape architecture. Some \cite{liu2018picanet,zhang2018progressive,wang2019salient} introduced various attention mechanisms to enhance the feature representation ability of the network.

In recent years, generative models have rapidly advanced and significantly influenced visual learning tasks, ranging from image synthesis~\cite{rombach2022high} to reinforcement learning~\cite{liu2025diffusion}. This trend has likewise motivated progress in image segmentation, where researchers have begun to integrate generative paradigms such as VAE-based approaches~\cite{zhang2021uncertainty}, GAN-driven frameworks~\cite{wang2019saliencygan}, and diffusion model-based techniques~\cite{sun2025conditional}. These methods leverage generative priors to refine feature representations and promote more stable and coherent segmentation results.

Compared with the above existing image segmentation methods, we perform a new cascading interaction mode of multi-scale information, combined with a global information guidance model, to reduce the loss of detailed information and improve accuracy.

\section{Method}\label{sec2}
To precisely segment the target and facilitate the visual servoing module in calculating its position, we use the Swin transformer~\cite{liu2021swin} as an encoder because of its unique advantages: the Swin transformer incorporates a local attention mechanism, inherits the advantages of CNNs in processing large images, and uses a window-based approach to exploit the transformer's capabilities in long-range dependency modeling. To extract scale-specific features based on different backbone networks, we introduce an additional convolutional layer with a kernel size of 1 to standardize the channel dimensions. Consequently, the resulting unified channel features can be denoted as $\mathcal{E} = \{E_i, 1\le i \le I\}$, where $I$ is typically set to 5. 

\begin{figure}[t]
    \centering
    \includegraphics[width=1\textwidth]{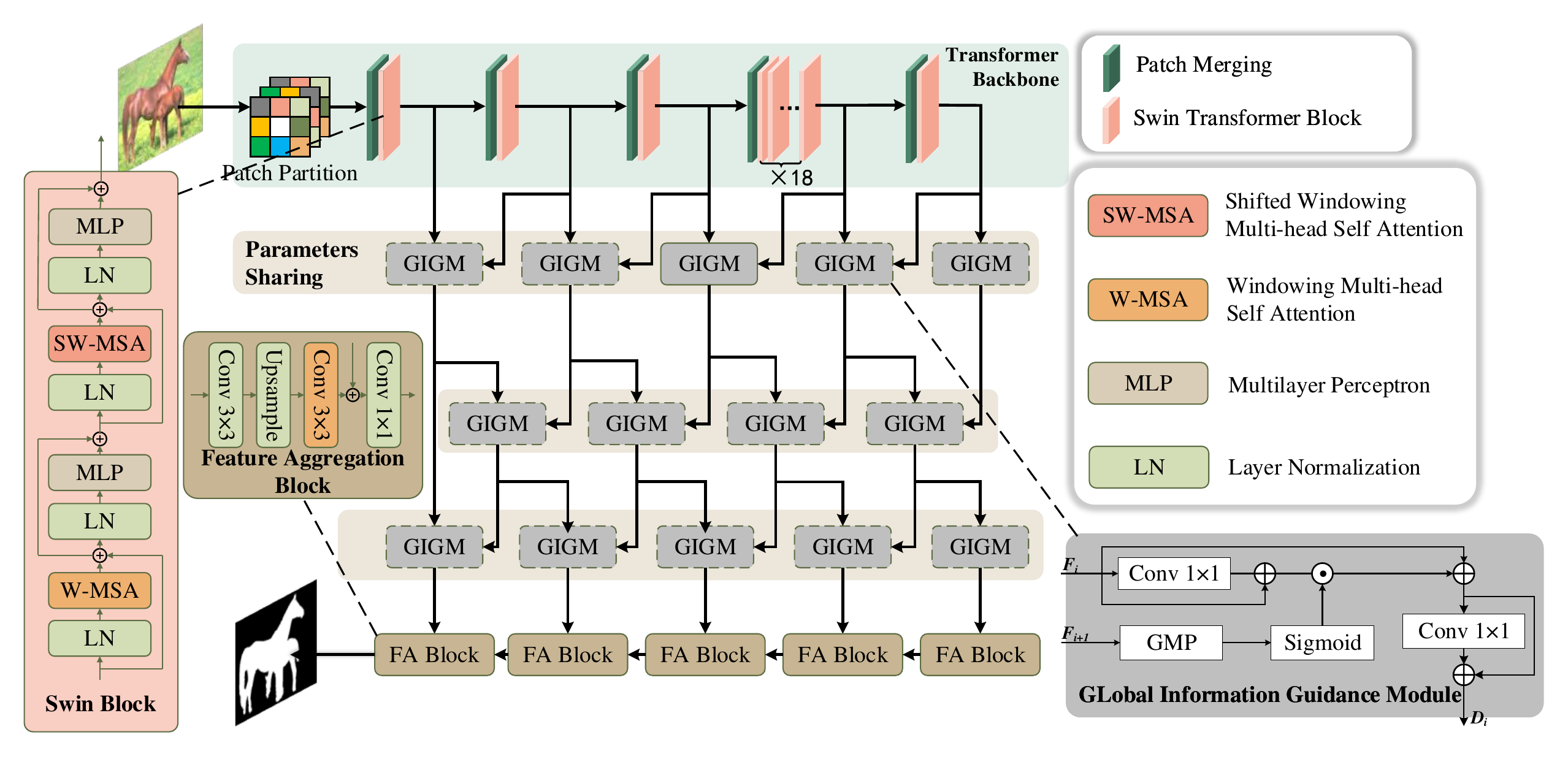}
    \caption{An overview of the proposed network framework}
    \label{wlkt}
\end{figure}

As shown in Fig. \ref{wlkt}, after applying convolutional pooling for down-sampling and subsequent up-sampling to restore the original resolution, images often suffer from blurring and loss of fine details. The conventional approach involves cascading feature maps at the same resolution along both the bottom-up and top-down paths, which mitigates the loss of local features to some extent. However, a direct feature extraction approach may limit multi-scale information fusion, as hierarchical feature interactions are often underutilized. To overcome this constraint, we propose a Cascaded Information Interaction Network, which enables multi-scale information exchange at the filter level. This technique establishes a structured mechanism for progressive feature refinement, ensuring effective communication across different resolution layers. Additionally, we recognize that deep architectures typically yield enhanced performance due to their ability to model complex patterns. Building on this idea, we expand the interaction layers in our model to strengthen hierarchical feature representation.
Given the channel unified feature maps from the encoder $\mathcal{E}$, the features delivered to the decoder $\mathcal{D} = \{D_i, 1\le j \le J\}$ could be got by cascaded interactors as

\begin{equation}
D_{j}=\mathbb{F}^{q}\left(E_{k}, \ldots, E_{m}\right), \quad 1 \leq j \leq 5, \quad 1\leq k \leq m \leq 5
\end{equation}
where $\mathbb{F}$ denotes the feature fusion in each interaction level, $q$ indicates the number of function actions, which means the number of cascading levels.

In segmentation tasks, an efficient multiscale module significantly enhances module performance. 
Higher-level information can serve to guide and enhance the interaction of lower-level information across different scales.
To maintain the compression of both local and relative global information, we introduce a global information guidance module (GIGM). 
The higher-level information can serve to guide the lower-level information, thereby enhancing the interaction between different scales of information. 
The module input contains the lower-level information $F_{i}$, which has been processed by a $1\times1$ convolutional layer. 
In addition, the higher-level information, $F_{i+1}$, has been subjected to Global Maximum Pooling (GMP) and sigmoid function, as shown by the gray box in Fig. \ref{wlkt}. 
The higher-level information is compressed to calibrate the lower-level information, thereby preserving local features. 
Finally, the output $D_i$ is obtained after a $1\times1$ convolutional layer. $D_i$ serves as an information guide from the relatively higher level pathway to the lower level pathway. The module is expressed as follows:
\begin{equation}
G_{i+1}=Sigmoid\left(G M P\left(F_{i+1}\right)\right), 1 \leq i \leq M-1
\end{equation}
\begin{equation} 
	D_i = (Conv^{1} + 1) (
		G_{i+1}
		\odot (Conv^{1} + 1 ) (F_i)+F_i), ~1\le i \le M. 
  \label{equ:rgc_overall}
\end{equation}

\section{Experimental Results}

\subsection{Experimental Setup}
The evaluation datasets utilized in our study include five well-established datasets: ECSSD ~\cite{yan2013hierarchical}, PASCAL-S ~\cite{li2014secrets}, DUT-OMRON ~\cite{yang2013saliency}, HKU-IS ~\cite{li2015visual}, and DUTS-TE ~\cite{wang2017learning}.
For model training, we consistently employ the DUTS-TR dataset ~\cite{wang2017learning} across all experiments, following established practices in image segmentation research.

Our model was trained for 60 rounds in batches of 30, and we selected the optimizer with a learning rate of 0.005, momentum of 0.9, and weight decay of 5e-5. The image input size was resized to 384 × 384 for both training and testing.
To assess the effectiveness of various methods, we utilize three commonly used metrics: the F-measure score ($F_\beta$), the mean absolute error ($MAE$), and the S-measure score ($S_\alpha$).
($F_\beta$) is calculated as follows:
\begin{equation}
	F_\beta = \frac{(1+\beta^2) \times \rm{Precision} \times \rm{Recall}}{\beta^2 \times \rm{Precision} + \rm{Recall} }.
\end{equation}

 To impose a higher weight for accuracy, we set $\beta^2$ to 0.3. At the pixel level, $MAE$ evaluates the average absolute difference between the predicted image $P$ and the labeled image $L$.
 \begin{equation}
	{\rm MAE} = \frac{1}{W \times H} \sum_{x=1}^{W} \sum_{y=1}^{H}|P(x, y)-L(x, y)|,
\end{equation}
where the width and height of the image are denoted by $W$ and $H$, respectively. The S-measure ($S_\alpha$) integrates both object-aware ($S_o$) and region-aware ($S_r$) structural similarity components, and is calculated as follows:
\begin{equation}
	S_\alpha = \gamma S_o + (1-\gamma)S_r,
\end{equation}
where $\gamma$ is 0.5 as is commonly done.

The loss function utilized in this paper combines an intersection-over-union (IoU) loss with a binary cross-entropy loss (BCE):
$l= l_{iou} + l_{bce}$.
Because of its excellent robustness, the binary cross-entropy (BCE) loss function is widely used in binary classification and is obtained by calculating the pixel-by-pixel loss of the image: 
\begin{equation}
l_{b c e}(p, l)=-\frac{1}{n} \sum_{k=1}^n\left[l_k \log \left(p_k\right)+\left(1-l_k\right) \log \left(1-p_k\right)\right]
\end{equation}
$p$ and $l$ stand for the predicted image and label, respectively. $k$ is the index of the pixel and $n$ is the number of pixels in $x$.
In contrast to the BCE loss function, which emphasizes differences at the pixel level, the IoU loss considers the overall graph similarity, and its definition is as follows:
\begin{equation}
l_{\text {iou }}(p, l)=1-\frac{\sum_{k=1}^n\left(l_k * p_k\right)}{\sum_{k=1}^n\left(l_k+p_k-l_k * p_k\right)} .
\end{equation}
\subsection{Comparisons to the State-of-the-Arts}
We compared the proposed image segmentation method with 22 state-of-the-art approaches, including  $\text{PAGR}$~\cite{zhang2018progressive}, $\text{DGRL}$~\cite{wang2018detect}, $\text{PiCANet}$~\cite{liu2018picanet}, $\text{MLMS}$~\cite{wu2019mutual}, $\text{PAGE}$~\cite{wang2019salient},  $\text{ICTB}$~\cite{wang2019iterative}, $\text{CPD}$~\cite{wu2019cascaded}, $\text{BASNet}$~\cite{qin2019basnet}, $\text{PoolNet}$~\cite{liu2019simple}, $\text{CSNet}$~\cite{gao2020sod100k},
$\text{GateNet}$~\cite{GateNet2020},
$\text{MINet}$~\cite{pang2020multi},
$\text{ITSD}$~\cite{zhou2020interactive},
$\text{VST}$~\cite{liu2021vst},
$\text{MSFNet}$~\cite{Miao_2021_ACM_MM},
$\text{CII}$~\cite{liu2021rethinking},
$\text{PoolNet+}$~\cite{liu2022poolnet+},
$\text{DCN}$~\cite{wu2021DCN}, 
$\text{DNA}$~\cite{yao2022BIPG}, 
$\text{RCSB}$~\cite{ke2022recursive}, 
$\text{PriorNet}$~\cite{zhu2024prior} and NASAL~\cite{liu2025towards}. 
To ensure a fair comparison, we either utilize saliency maps shared by the authors or compute their released models. We then quantitatively compare the obtained results by calculating the F-measure score $F_\beta$, the S-measure score $S_\alpha$ and the mean absolute error (MAE) of our method alongside the other methods. 
Table \ref{t2} presents the results of the other advanced measurement methods mentioned.
On the ECSSD dataset, our method achieves the highest $F_\beta$ (0.952) and the lowest MAE (0.028), while maintaining a high $S_\alpha$ value of 0.933. These results suggest enhanced capacity for capturing fine details and complex object structures, particularly in cluttered scenes. Similarly, on PASCAL-S, our model maintains leading performance, with minimized MAE and competitive $F_\beta$ and $S_\alpha$ values, indicating improved robustness in handling occlusion and challenging backgrounds.
Performance on HKU-IS further highlights the model’s generalization capabilities, recording an $F_\beta$ of 0.898, MAE of 0.031, and $S_\alpha$ of 0.929, surpassing comparative methods across all metrics. On more challenging datasets such as DUT-OMRON and DUTS-TE, the method maintains its advantages, our method shows significant improvement of $1.1\%$ and $0.8\%$ compared with the famous PoolNet+ model~\cite{liu2022poolnet+}, which confirms its effectiveness in delineating object boundaries under complex scenes.
Our model achieves leading performance in salient object detection, owing to its unique architecture that combines multi-scale feature interaction with global information guidance. This design enhances detail preservation while maintaining accurate global context.
\begin{sidewaystable}[htbp]
\renewcommand{\arraystretch}{1.15}
    \renewcommand{\tabcolsep}{0.01mm}
\caption{Comparisons of our method with other state-of-the-art methods on five popular SOD benchmarks.}\label{t2}
\begin{tabular*}{\textheight}{@{\extracolsep\fill}lccccccccccccccc}
\toprule%
\multirow{2}*{Method}  &
        \multicolumn{3}{c}{ECSSD } & \multicolumn{3}{c}{PASCAL-S } & \multicolumn{3}{c}{DUT-OMRON} & \multicolumn{3}{c}{HKU-IS} & \multicolumn{3}{c}{DUTS-TE}
        \\
    \cmidrule{2-4}
    \cmidrule{5-7}
    \cmidrule{8-10}
    \cmidrule{11-13}
    \cmidrule{14-16}
    & $F_\beta$~$\uparrow$ & MAE~$\downarrow$ &$S_\alpha$$\uparrow$ & $F_\beta$~$\uparrow$ & MAE~$\downarrow$ &$S_\alpha$$\uparrow$ & $F_\beta$~$\uparrow$ & MAE~$\downarrow$ &$S_\alpha$$\uparrow$ & $F_\beta$~$\uparrow$ & MAE~$\downarrow$ &$S_\alpha$$\uparrow$& $F_\beta$~$\uparrow$ & MAE~$\downarrow$&$S_\alpha$$\uparrow$\\

\midrule
        $\text{PAGR}$~\cite{zhang2018progressive} & 0.927 & 0.061 & 0.889 & 0.847 & 0.089 & 0.822 & 0.771 & 0.071 & 0.775 & 0.919 & 0.047 & 0.889  & 0.854 & 0.055 & 0.839 \\
        $\text{DGRL}$~\cite{wang2018detect} & 0.922 & 0.041 & 0.903 & 0.844 & 0.072 & 0.836 & 0.774 & 0.062 & 0.806 & 0.910 & 0.036 & 0.895 & 0.828 & 0.049 & 0.842 \\
        $\text{PiCANet}$~\cite{liu2018picanet} & 0.935 & 0.047 & 0.917 & 0.864 & 0.075 & 0.854 & 0.820 & 0.064 & 0.830 & 0.920 & 0.044 & 0.904 & 0.863 & 0.050 & 0.868 \\
        $\text{MLMS}$~\cite{wu2019mutual} & 0.930 & 0.045 & 0.911 & 0.853 & 0.074 & 0.844 & 0.793 & 0.063 & 0.809 & 0.922 & 0.039 & 0.907 & 0.854 & 0.048 & 0.862 \\ 
        $\text{PAGE}$~\cite{wang2019salient} & 0.931 & 0.042 & 0.912 & 0.848 & 0.076 & 0.842 & 0.791 & 0.062 & 0.825 & 0.920 & 0.036 & 0.904  & 0.838 & 0.051 & 0.855 \\ 
        
        $\text{ICTB}$~\cite{wang2019iterative} & 0.938 & 0.041 & 0.918 & 0.855 & 0.071 & 0.850 & 0.811 & 0.060 & 0.837 & 0.925 & 0.037 & 0.909 & 0.855 & 0.043 & 0.865 \\
        $\text{CPD}$~\cite{wu2019cascaded} & 0.939 & 0.037 & 0.918 & 0.859 & 0.071 & 0.848 & 0.796 & 0.056 & 0.825 & 0.925 & 0.034 & 0.907 & 0.865 & 0.043 & 0.869 \\ 
        $\text{BASNet}$~\cite{qin2019basnet} & 0.942 & 0.037 & 0.916 & 0.857 & 0.076 & 0.838 & 0.811 & 0.057 & 0.836 & 0.930 & 0.033 & 0.908 & 0.860 & 0.047 & 0.866 \\ 
        $\text{PoolNet}$~\cite{liu2019simple} & 0.944 & 0.039 & 0.921 & 0.865 & 0.075 & 0.850 & 0.830 & 0.055 & 0.836 & 0.934 & 0.032 & 0.917 & 0.886 & 0.040 & 0.883 \\
        $\text{CSNet}$~\cite{gao2020sod100k} & 0.944 & 0.038 & 0.921 & 0.866 & 0.073 & 0.851 & 0.821 & 0.055 & 0.831 & 0.930 & 0.033 & 0.911 & 0.881 & 0.040 & 0.879 \\
        $\text{GateNet}$~\cite{GateNet2020} & 0.946 & 0.040 & 0.920 & 0.877 & 0.068 & 0.858 & 0.831 & 0.055 & 0.838 & 0.935 & 0.033 & 0.915 & 0.889 & 0.040 & 0.885 \\ 
        $\text{MINet}$~\cite{pang2020multi}  & 0.947 & 0.034 & 0.925 & 0.874 & 0.064 & 0.856 & 0.826 & 0.056 & 0.833 & 0.936 & 0.028 & 0.920 & 0.888 & 0.037 & 0.884 \\
        $\text{ITSD}$~\cite{zhou2020interactive} & 0.947 & 0.035 & 0.925 & 0.871 & 0.066 & 0.859 & 0.823 & 0.061 & 0.840 & 0.933 & 0.031 & 0.916 & 0.883 & 0.041 & 0.885\\
        $\text{VST}$~\cite{liu2021vst}  & 0.951 & 0.034 &0.932 & 0.875 & 0.062 & 0.872& 0.829 & 0.058 & 0.850& 0.942 & 0.030 & \textbf{0.929}& 0.891 & 0.037 & 0.896\\
        % $\text{MSFNet}$~\cite{Miao_2021_ACM_MM}  & 0.943 & 0.033 & & 0.865 & 0.061 & & 0.824 & 0.050 & & 0.930 & 0.027 & & 0.881 & 0.034 & \\
        $\text{MSFNet}$~\cite{Miao_2021_ACM_MM}  & 0.943 & 0.033 & 0.915& 0.865 & 0.061 &0.852& 0.824 & 0.050 &0.832& 0.930 & 0.027 &0.909& 0.881 & 0.034 &0.877\\
        % $\text{PurNet}$~\cite{PurNet}  & ,, & ,, & ,, & ,, & ,, & ,, & ,, & ,, & ,, & ,, \\
        % \hline
        $\text{CII}$~\cite{liu2021rethinking}  & 0.950 & 0.034 & 0.926 & 0.882 & 0.062 & 0.865 & 0.831 & 0.054 & 0.839 & 0.939 & 0.029 & 0.920 & 0.890 & 0.036 & 0.888\\ 
        $\text{PoolNet+}$~\cite{liu2022poolnet+} & 0.949 & 0.040 & 0.925 & 0.879 & 0.068 & 0.864 & 0.831 & 0.056 & 0.842 & 0.941 & 0.034 & 0.921 & 0.894 & 0.039 & 0.890 \\
        $\text{DCN}$~\cite{wu2021DCN}  & 0.952 & 0.031 &0.928 & 0.872 & 0.062 & 0.861& 0.823 & 0.051 & 0.845& 0.940 & 0.027 & 0.922& 0.894 & 0.035 & 0.891\\
        $\text{DNA}$~\cite{DNA2022}  & 0.940 & 0.043 &0.915& 0.855  & 0.079 & 0.837& 0.803  & 0.063 & 0.818& 0.927  & 0.036 &0.905&  0.873 & 0.046 & 0.860\\
        $\text{RCSB}$~\cite{ke2022recursive}  & 0.945 & 0.033 & 0.922& 0.879  & 0.059 & 0.860& \textbf{0.849}  & \textbf{0.049} & 0.835& 0.939  & 0.027 &0.918&  0.897 & 0.035 &0.881\\
        $\text{PriorNet}$~\cite{zhu2024prior}  & \textbf{0.953} & 0.031 &0.931& 0.881  & 0.059 &0.869& 0.839  & 0.051 &0.849 & 0.940  & 0.029 &0.920& \textbf{0.901} & 0.033 &0.897\\
        $\text{NASAL}$~\cite{liu2025towards} & 0.925 & 0.052 & 0.904 & 0.836 & 0.092 & 0.825 & 0.800 & 0.069 & 0.818 & 0.913 & 0.044 & 0.898 & 0.833 & 0.060 & 0.841 \\
        \midrule
        $\textbf{Ours}$ & 0.952 & \textbf{0.028}&\textbf{0.933} & \textbf{0.888} & \textbf{0.054} &\textbf{0.879} & 0.842 & \textbf{0.049} & \textbf{0.858}& \textbf{0.943} & \textbf{0.025} &\textbf{0.929} & 0.898 & \textbf{0.031} &\textbf{0.900}\\
\botrule
\end{tabular*}
\end{sidewaystable}

\begin{figure}[htbp]
\centering
 \begin{minipage}{0.11\linewidth}
 	\vspace{3pt}
 	\centerline{\includegraphics[width=\textwidth]{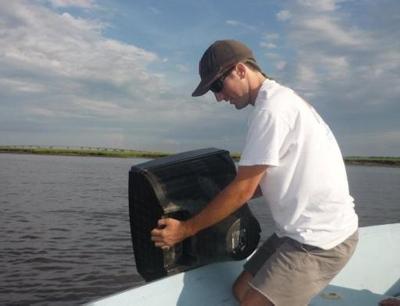}}
 	\vspace{3pt}
 	\centerline{\includegraphics[width=\textwidth]{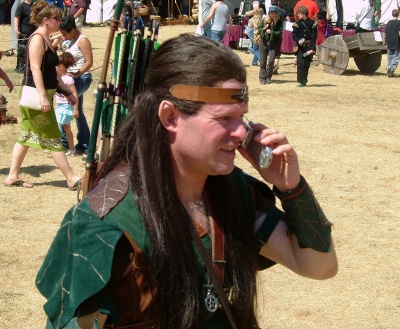}}
 	\vspace{3pt}
 	\centerline{\includegraphics[width=\textwidth]{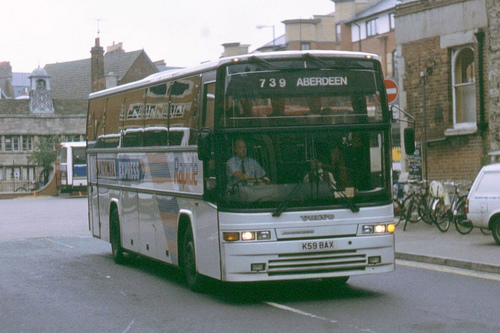}}
 	\vspace{3pt}
 	\centerline{\includegraphics[width=\textwidth]{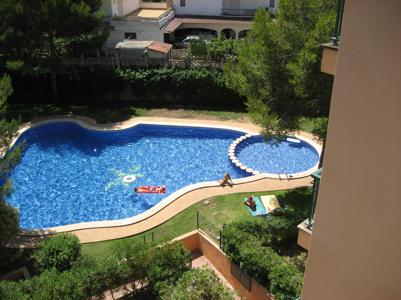}}
 	\vspace{3pt}
 	\centerline{\includegraphics[width=\textwidth]{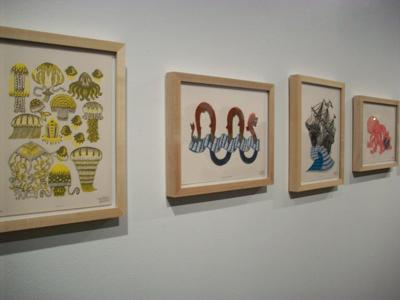}}
 	\vspace{3pt}
 \end{minipage}
 \begin{minipage}{0.11\linewidth}
	\vspace{3pt}
	\centerline{\includegraphics[width=\textwidth]{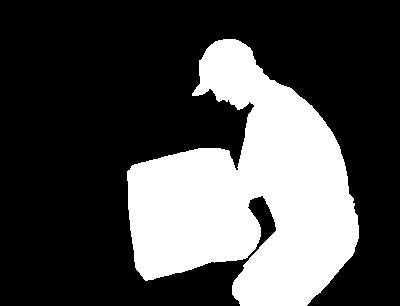}}
	\vspace{3pt}
 	\centerline{\includegraphics[width=\textwidth]{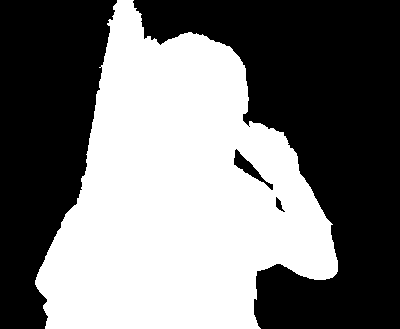}}
	\vspace{3pt}
	\centerline{\includegraphics[width=\textwidth]{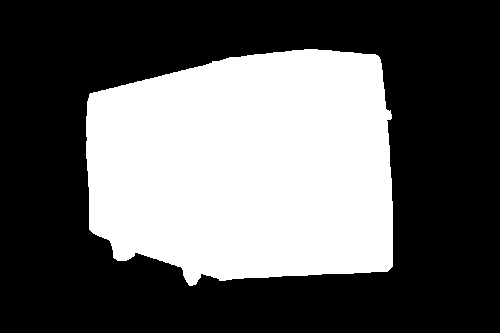}}
	\vspace{3pt}
	\centerline{\includegraphics[width=\textwidth]{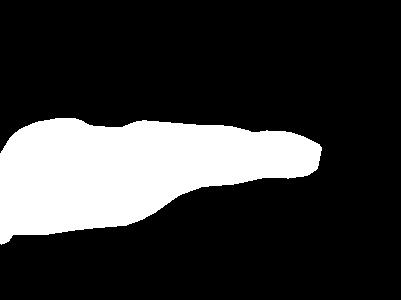}}
	\vspace{3pt}
	\centerline{\includegraphics[width=\textwidth]{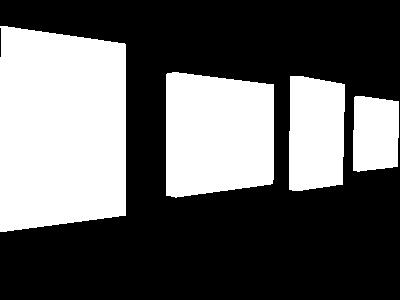}}
	\vspace{3pt}
\end{minipage}
\begin{minipage}{0.11\linewidth}
	\vspace{3pt}
	\centerline{\includegraphics[width=\textwidth]{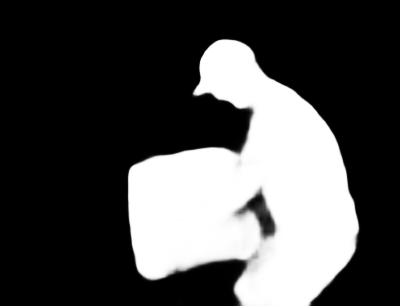}}
	\vspace{3pt}
	\centerline{\includegraphics[width=\textwidth]{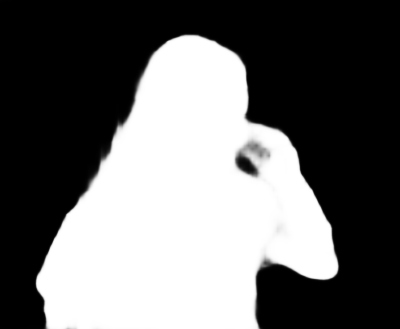}}
	\vspace{3pt}
	\centerline{\includegraphics[width=\textwidth]{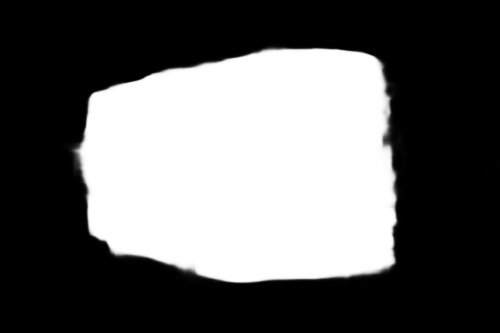}}
	\vspace{3pt}
	\centerline{\includegraphics[width=\textwidth]{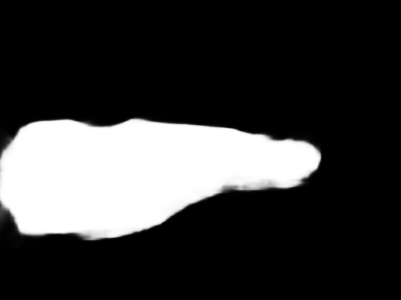}}
	\vspace{3pt}
	\centerline{\includegraphics[width=\textwidth]{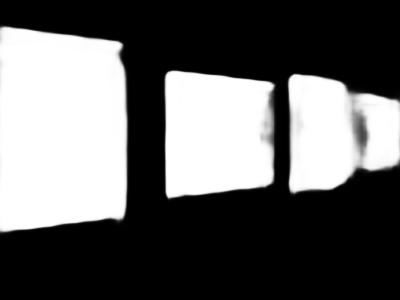}}
	\vspace{3pt}
\end{minipage}
 \begin{minipage}{0.11\linewidth}
 	\vspace{3pt}
 	\centerline{\includegraphics[width=\textwidth]{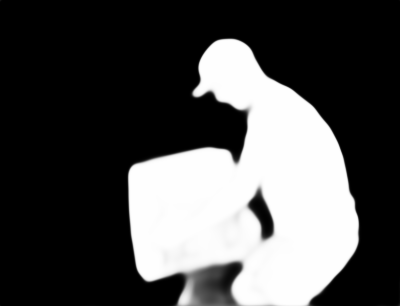}}
 	\vspace{3pt}
 	\centerline{\includegraphics[width=\textwidth]{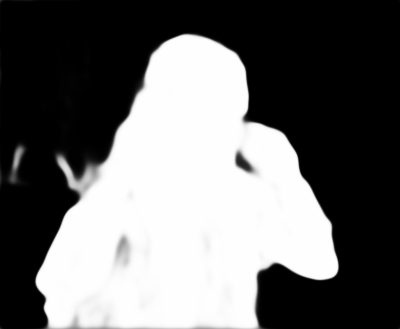}}
 	\vspace{3pt}
 	\centerline{\includegraphics[width=\textwidth]{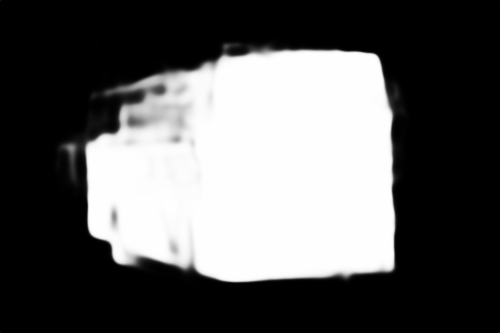}}
 	\vspace{3pt}
 	\centerline{\includegraphics[width=\textwidth]{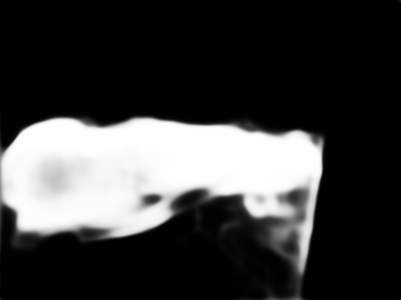}}
 	\vspace{3pt}
 	\centerline{\includegraphics[width=\textwidth]{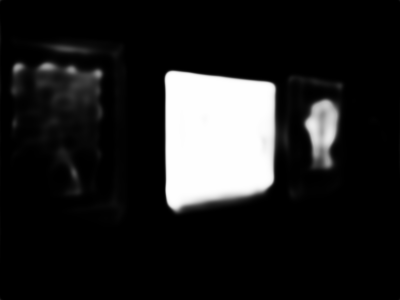}}
 	\vspace{3pt}
 \end{minipage}
 \begin{minipage}{0.11\linewidth}
	\vspace{3pt}
	\centerline{\includegraphics[width=\textwidth]{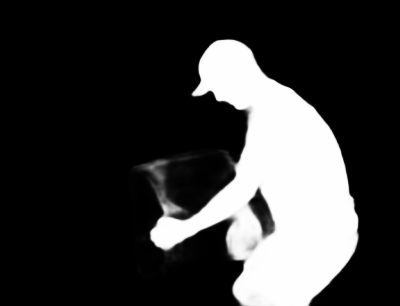}}
	\vspace{3pt}
 	\centerline{\includegraphics[width=\textwidth]{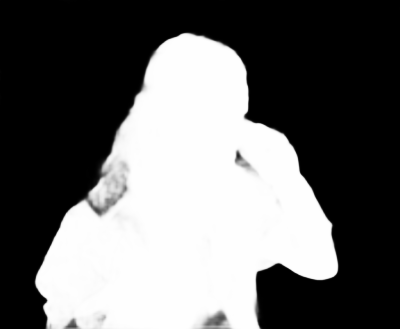}}
	\vspace{3pt}
	\centerline{\includegraphics[width=\textwidth]{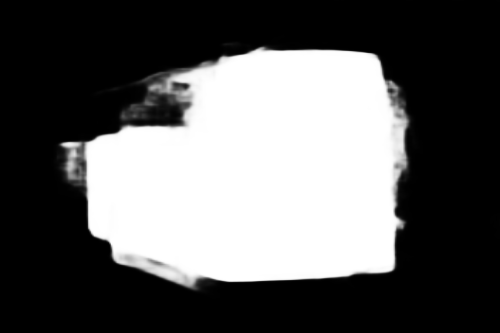}}
	\vspace{3pt}
	\centerline{\includegraphics[width=\textwidth]{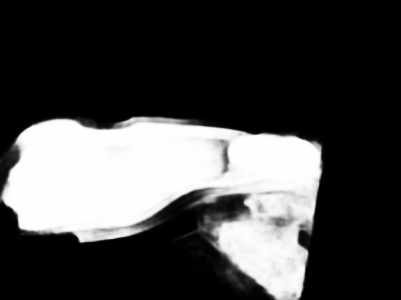}}
	\vspace{3pt}
	\centerline{\includegraphics[width=\textwidth]{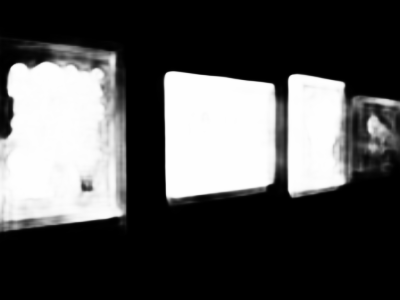}}
	\vspace{3pt}
\end{minipage}
\begin{minipage}{0.11\linewidth}
	\vspace{3pt}
	\centerline{\includegraphics[width=\textwidth]{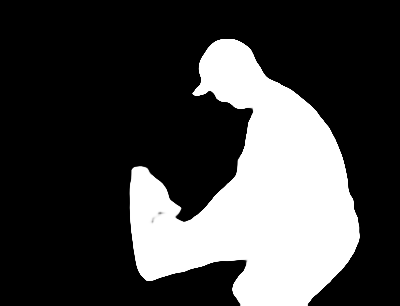}}
	\vspace{3pt}
	\centerline{\includegraphics[width=\textwidth]{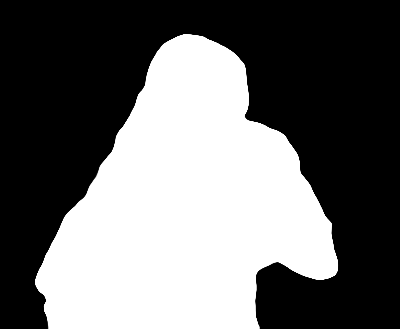}}
	\vspace{3pt}
	\centerline{\includegraphics[width=\textwidth]{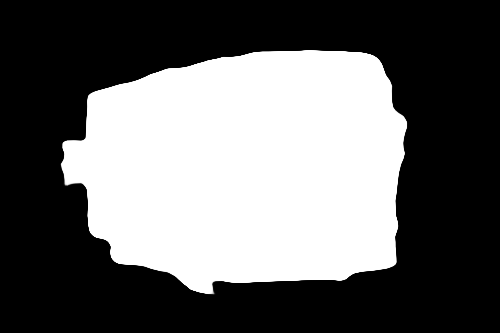}}
	\vspace{3pt}
	\centerline{\includegraphics[width=\textwidth]{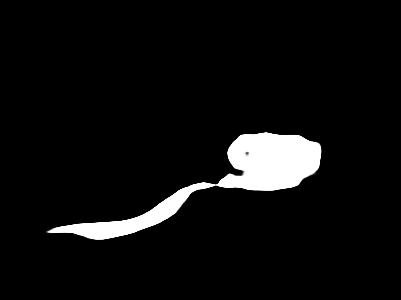}}
	\vspace{3pt}
	\centerline{\includegraphics[width=\textwidth]{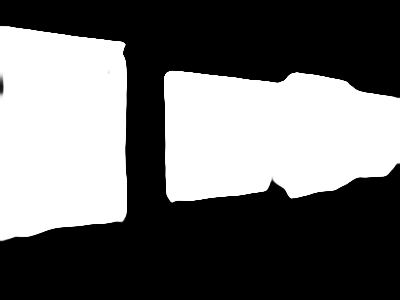}}
	\vspace{3pt}
\end{minipage}
\begin{minipage}{0.11\linewidth}
	\vspace{3pt}
	\centerline{\includegraphics[width=\textwidth]{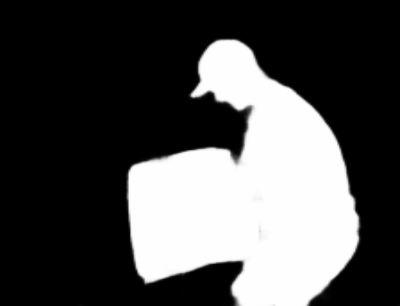}}
	\vspace{3pt}
	\centerline{\includegraphics[width=\textwidth]{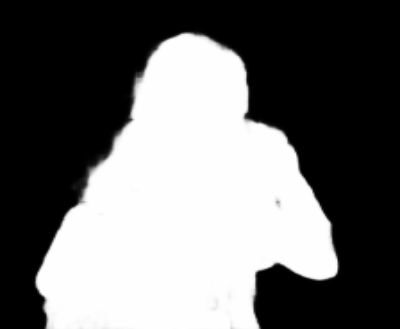}}
	\vspace{3pt}
	\centerline{\includegraphics[width=\textwidth]{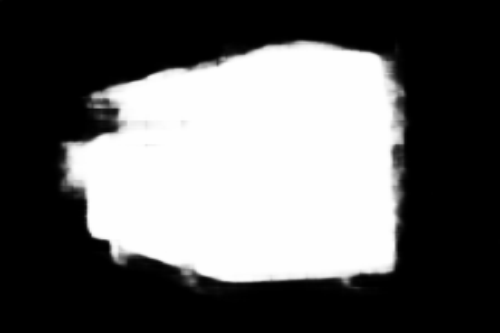}}
	\vspace{3pt}
	\centerline{\includegraphics[width=\textwidth]{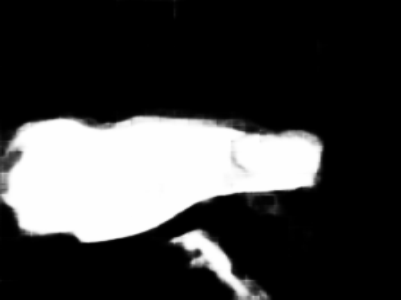}}
	\vspace{3pt}
	\centerline{\includegraphics[width=\textwidth]{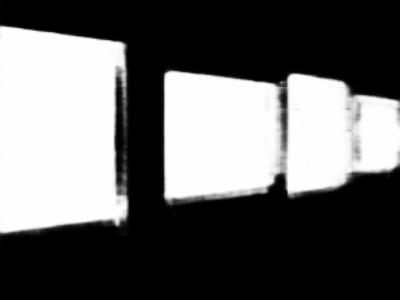}}
	\vspace{3pt}
\end{minipage}
\begin{minipage}{0.11\linewidth}
 	\vspace{3pt}
 	\centerline{\includegraphics[width=\textwidth]{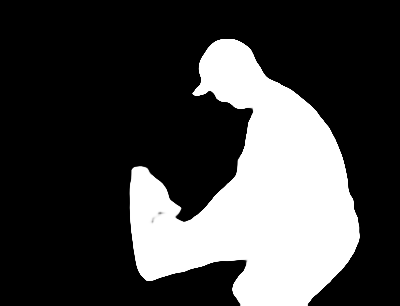}}
 	\vspace{3pt}
 	\centerline{\includegraphics[width=\textwidth]{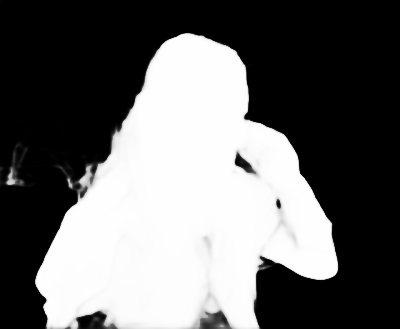}}
 	\vspace{3pt}
 	\centerline{\includegraphics[width=\textwidth]{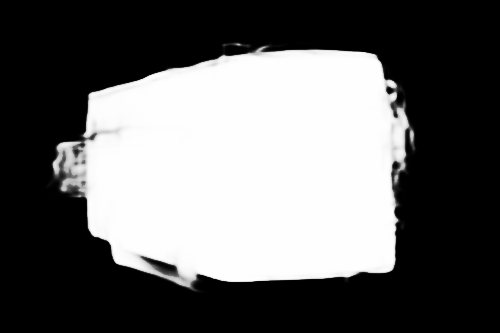}}
 	\vspace{3pt}
 	\centerline{\includegraphics[width=\textwidth]{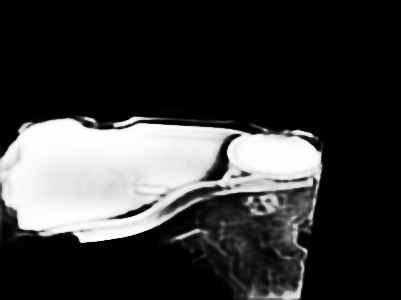}}
 	\vspace{3pt}
 	\centerline{\includegraphics[width=\textwidth]{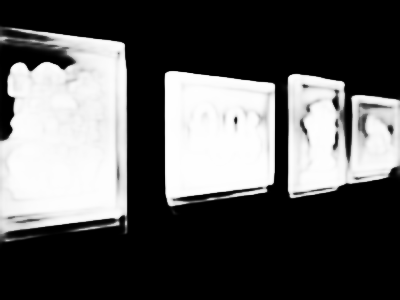}}
 	\vspace{3pt}
 \end{minipage}
\caption{Visual comparison of saliency maps with state-of-the-art methods. From left to right: Input image, Ground truth, Ours, DNA, CII, MSFNet, VST and ITSD. Our approach consistently produces the best results. }
\label{sotafig}
\end{figure}

In Fig. \ref{sotafig}, we present example saliency maps generated by our method. These maps demonstrate our method’s ability to produce accurate results with clear boundaries and uniform highlights. 

\section{Conclusion}

In this work, we presented a Cascade Interaction Network designed to enhance information interaction capabilities and improve the robustness of image segmentation models. Central to our approach is the integration of a Global Information Guidance Module, which facilitates the effective fusion of low-level texture details and high-level semantic features. This mechanism successfully mitigates the limitations of single-scale feature extraction, ensuring high segmentation accuracy even in visually cluttered or blurred environments. Extensive experiments and comparisons on standard datasets verify that our proposed framework not only outperforms existing methods in terms of precision but also maintains the efficiency required for practical deployment. These results suggest that our model is a promising solution for visual perception in autonomous robotic systems.

\bibliography{sn-bibliography}% common bib file

\end{document}